\documentclass[11pt]{article}

\usepackage[left=1.25in,top=1in,right=1.25in,bottom=1in,nohead]{geometry}
\usepackage{graphicx}
\usepackage{amsfonts,amssymb,amsmath,amsthm}
\usepackage[numbers,square]{natbib}

\usepackage{hyperref}
\usepackage{pdfsync}

\usepackage{algorithm,algorithmic}

\newcommand{\email}[1]{\href{mailto:#1}{#1}}

\newcommand{\affinity}{W}
\newcommand{\tune}{\epsilon}

\newcommand{\nys}{Nystr\"om }
\newcommand{\nysNoSpace}{Nystr\"om}
\newcommand{\approxEig}{Gaussian projection }
\newcommand{\approxEigNoSpace}{Gaussian projection}
\newcommand{\ntrain}{n_{\textrm{train}}}
\newcommand{\ntest}{n_{\textrm{test}}}
\newcommand{\E}{\mathbb{E}}

\title{Spectral approximations in machine learning}

\author{
Darren Homrighausen \\
Department of Statistics\\
Carnegie Mellon University\\
Pittsburgh, PA 15213 \\
\email{dhomrigh@stat.cmu.edu} \\
\and
Daniel J. McDonald\\
Department of Statistics\\
Carnegie Mellon University\\
Pittsburgh, PA 15213 \\
\email{danielmc@stat.cmu.edu} \\
}

\date{Version: July 20, 2011}
\begin{document}

\maketitle

\begin{abstract}
In many areas of machine learning, it becomes necessary to find the
eigenvector decompositions of large matrices. We discuss two methods for reducing the
computational burden of spectral decompositions: the more venerable
\nys extension and a newly introduced algorithm based on random
projections. Previous
work has centered on the ability to reconstruct the
original matrix. We argue that a more interesting and
relevant
comparison is their relative performance in clustering and classification tasks using the
approximate eigenvectors as features. We demonstrate that performance is
task specific and depends on the rank of the approximation.
\end{abstract}

\section{Introduction}
\label{sec:sca}
Spectral Connectivity Analysis (SCA) is 
a burgeoning area in statistical machine learning. 
Beginning with principal
components analysis (PCA) 
and Fisher discriminant analysis 
and continuing with locally linear embeddings \citep{roweis2000},
SCA techniques for discovering low-dimensional metric embeddings of the data
have long been a part of data analysis.
Moreover, newer techniques, such as Laplacian eigenmaps \citep{belkin2003}
and Hessian maps \citep{donoho2003}, similarly seek to elucidate
the underlying geometry of the data by analyzing approximations to certain operators
and their respective eigenfunctions. 

Though the techniques are different in many respects, they
share one major characteristic: the necessity for the computation of a postive definite
kernel and, more importantly, its spectrum. Unfortunately,
a well known speed limit exists. For a matrix $\affinity \in \mathbb{R}^{n\times n}$,
the eigenvalue decomposition can be computed no faster than $O(n^3)$.
On its face, this limit constrains the applicability of SCA techniques to moderately 
sized problems.

Fortunately, there exist methods in numerical 
linear algebra for the approximate computation of the spectrum of a
matrix.  Two such methods are the \nys extension~\citep{WilliamsSeeger2001}
and another algorithm ~\citep{HalkoMartinsson2009} we have dubbed `\approxEigNoSpace' 
for reasons that will become clear.  Each seeks to compute the eigenvectors of a 
smaller matrix 
and then embed the remaining structure of the original matrix onto these eigenvectors.  
These methods are widely
used in the applied SCA community e.g. \citep{freeman2009,fowlkes2004}.
However, the choice as to the approximation method has remained a subjective one,
largely based on the subfield of the respective researchers. 

The goal of this paper is to provide a careful, comprehensive comparison
of the \nys extension and \approxEig methods for approximating the spectrum of
the large matrices occuring in SCA techniques.  As SCA is such a vast field, we concentrate
on one specific technique, known as diffusion map from the graph Laplacian.
In this technique, it is necessary to compute the spectrum of the Laplacian matrix $\mathbb{L}$.  
For a dataset with $n$ observations, 
the matrix $\mathbb{L}$ is $n$ by $n$, and hence the exact numerical spectrum quickly becomes
impossible to compute. 

Previous work on a principled comparison of the \nys extension and \approxEig methods 
have left open some important
questions.  For instance, \cite{drineas2005,TalwalkarKumar2008} both provide contrasts
that, while interesting, are limited in that they only consider
approximation methods based on column sampling. 
Furthermore, performance evaluations in the
literature focus on reconstructing the matrix $\affinity$. However, in the
machine learning community, this is rarely the appropriate metric. Generally,
some or all of the eigenvectors are used as an input to standard
classification, regression, or clustering algorithms. Hence, there is
a need for comparing the effect of these methods on the accuracy of learning machines.

We find that for these sorts of applications, neither \nys nor
\approxEig achieves uniformly better results. For similar computational costs,
both methods perform well at manifold reconstruction. However, \approxEig
performs relatively poorly in a related clustering task.  Yet, it also outperforms \nys
on a standard classification task. We also find
that the choice of tuning parameters can be easier for one method
relative to the other depending on the application. 

The remainder of this paper is structured as follows.  In \S\ref{sec:nystromVSapprox} we 
give an overview of our
particular choice of SCA technique --- the
diffusion map from the graph
Laplacian --- as well as introduce both the \nys
and the \approxEig methods. In
\S\ref{sec:experimental-results}, we present our empirical
results on three tasks: low dimensional manifold recovery, clustering, and classification. 
Lastly, \S\ref{sec:discussion} details areas in need of further
investigation and provides a synthesis of our findings about the strengths of each
approximation method.

\section{Diffusion maps and spectral approximations}
\label{sec:nystromVSapprox}

\subsection{Diffusion map with the graph Laplacian}
\label{sec:diffusionMap}
The technique in SCA that we consider is commonly referred to as a diffusion map 
from the graph Laplacian.  Roughly, the idea is to construct an adjacency graph on
a given data set and then find a parameterization for the data which minimizes an objective
function that preserves locality.  We give a brief overview here.  See
\citep{lee2010} for a more comprehensive treatment. 

Specifically, suppose we
have $n$ observations, $x_1,\ldots,x_n$.  Define a graph $G=(V,E,\widetilde W)$ 
on the data such that $V=\{x_1,\ldots,x_n\}$
are the observations, $E$ is the set of connections between pairs of observations,
and $\widetilde W$ is a weight matrix associated with every element in $E$, corresponding
to the strength of the edge.
We make the common choice (e.g.~\cite[\S2.2.1]{lee2008}) of defining 
\begin{equation}
\widetilde{W}_{ij} = \exp\left\{-||x_i-x_j||^2/\tune\right\}
\label{eq:kernelDef}
\end{equation} 
for all $1 \leq i,j \leq n$
such that $(i,j) \in E$.  Further, we define a normalized version of the matrix $W$.
This can be done either symmetrically, as 
\begin{equation}
W := D^{-1/2} \widetilde W D^{-1/2},
\label{eq:symmetric}
\end{equation}
or asymmetrically as 
\begin{equation}
W := D^{-1} \widetilde W,
\label{eq:asymmetric}
\end{equation}
where $D :=$ diag$(\sum_{j=1}^n \widetilde{W}_{1j},\ldots,\sum_{j=1}^n \widetilde{W}_{nj})$.
In either case, the graph Laplacian is defined to be
$\mathbb{L} := I - W$.  Here, $I$ is the $n$ by $n$ identity matrix.

One can show (\cite[\S8.1]{lee2008}) that the optimal $p$
dimensional embedding is given by the diffusion map onto
eigenvectors two through $p+1$ of $\affinity$.\footnote{The first eigenvalue and
associated eigenvector correspond to the trivial solution and are 
discarded.}  That is, find an orthonormal matrix $U$
and diagonal matrix $\Sigma$ such that $\affinity = U \Sigma U^{\top}$
and retain the second through $(p+1)^{th}$ column of $U$ as the features.
Note that we have supressed any mention of eigenvalues in the diffusion map as
they only represent rescaling in our applications and hence are irrelevant to
our conclusions. Also, $\mathbb{L}$ and $\affinity$ have the same eigenvectors.

In most machine learning scenarios, this map is created as the first
step for more traditional applications.
It can be used
as a design (also known as a feature) matrix in classification or regression.  Also,
it can be used as coordinates of the data for unsupervised (clustering) techniques.

The crucial aspect is the need to compute the eigenvectors of the matrix $\affinity$.
We give an overview of the \nys method in \S\ref{sec:nystrom} 
and \approxEig method in \S\ref{sec:approx} for approximating these eigenvectors.

\subsection{Nystr\"om}
\label{sec:nystrom}
As mentioned in \S\ref{sec:sca}, for most cases of interest in machine learning,
$n$ is large and hence the computation of the eigenvectors of the matrix
$\affinity$ is computationally infeasible. However, the \nys
extension gives a method for computing the eigenvectors and eigenvalues
of a smaller matrix, formed by column subsampling, 
and `extending' them to the remaining columns.

The \nys method for approximating the eigenvectors of a matrix 
comes from a much older technique for finding a numerical solution of
integral equations.  For our purposes, the \nys method finds
eigenvectors of a reduced
matrix that approximates the action of $\affinity$.\footnote{We
  consider the numerical spectral decomposition of the full matrix
  $\affinity$ to be the `exact' or `true' decomposition and ignore the effects of
numerical error in our nomenclature.}  Specifically, choose 
an integer $m < n$ and an associated index subset $M \subset \{1,\ldots,n\} =: N$ such that
$|M| = m$ to form an $m \times m$ matrix $\affinity^{(m)} := \affinity_{M,M}$.  Here
our notation mimics common coding syntax. Subsetting a matrix with a set of indices 
indicates the retention of those rows or columns.  Then,
we find a diagonal matrix $\Lambda^{(m)}$ and orthonormal matrix $U^{(m)}$ such that
\begin{equation}
\label{eq:nystrom}
\affinity^{(m)}U^{(m)} = \Lambda^{(m)}U^{(m)}.
\end{equation}
These eigenvectors are then extended to an approximation of the matrix $U$
by a simple formula.  See Algorithm \ref{alg:nystrom} 
for details and an outline of the procedure.  

Of course, choosing the subset $M$ is important.  By far the most common technique is
to choose $M$ by sampling $m$ times without replacement from $N$.  We
refer to this as the `uniform Nystr\"om' method.  Recently, however, there has been work on
making more informed and data-dependent choices.  The first work in this area
we are aware of can be found in \cite{drineas2005}.  More recently, \cite{BelabbasWolfe2009}
proposed sampling from a distribution formed by all $\binom{n}{m}$ determinants of the possible
retained matrices.  This is of course not feasible, and \cite{BelabbasWolfe2009} provides
some approximations.  We use the scheme where instead of a uniform draw from $N$,
we draw from $N$ in proportion to the size of the diagonal elements in $\affinity$.  When used,
this method is referred to as the `weighted Nystr\"om.'

\begin{algorithm}[t]
\caption{Nystr\"om approximation based on subsampling} 
Given an $n \times n$ matrix $\affinity$ and integer $m<n$, compute
an approximation $U^{nys}$ to the eigenvectors $U$ of $\affinity$. 
\begin{algorithmic}[1]
\STATE Compute $U^{(m)}$ and $\Lambda^{(m)}$
via equation \eqref{eq:nystrom}.
\STATE Form
\begin{align*}
  \widehat{\lambda}_i &= \frac{n}{m} \lambda_i^{(m)} 
  &&\widehat{\mathbf{u}}_i = \sqrt{\frac{m}{n}} \frac{1}{\lambda_i^{(m)}} \affinity_{N,M} \mathbf{u}_i^{(m)},
\end{align*}
where $\lambda_i^{(m)}$ and $\mathbf{u}_i^{(m)}$ are the $i^{th}$ diagonal entry and $i^{th}$ column
of $\Lambda^{(m)}$ and $U^{(m)}$, respectively.
\STATE Return $U^{nys} = [\widehat{\mathbf{u}}_1, \ldots, \widehat{\mathbf{u}}_m]$
\end{algorithmic} 
\label{alg:nystrom} 
\end{algorithm}

\subsection{Gaussian projection}
\label{sec:approx}
An alternate to the \nys method is
a very interesting new algorithm introduced in~\citep{HalkoMartinsson2009}.
This new algorithm differs from other
approximation techniques in that it produces, for any
matrix $A$, a subspace of col$(A)$ (the column space of $A$) through the action of $A$ on
a random set of linearly independent vectors.  This is in contrast to randomly
selecting columns of $A$ to form a subspace of col$(A)$ as in \citep{drineas2005}
and the \nys method.
This is important as crucial features of col$(A)$ can be missed by
simply subsampling columns.

We include a combination of algorithms 4.1 and 5.3 
from \citep{HalkoMartinsson2009} in Algorithm \ref{alg:approx}
 for completeness.

\subsection{Theoretical comparisons}
\label{sec:theoretical-results}

For both the \nys and the \approxEig methods, theoretical results
have been developed demonstrating their performance at forming a 
rank $m$ approximation to $\affinity$. For the weighted \nys method,
\citep{BelabbasWolfe2009} finds that 
\begin{equation}
  \E||\affinity - \affinity^{nys}||_F \leq ||\affinity - \affinity_{m}|| 
  + C \sum_{i=1}^n \affinity_{ii}^2
  \label{eq:nystromTheory}
\end{equation}
where $\affinity^{nys}$ is the approximation to $\affinity$ via the \nys method
and $\affinity_m$ is the best rank $m$ approximation.  Likewise,
for \approxEig, \citep{HalkoMartinsson2009} finds that
\begin{equation}
  ||\affinity - \affinity^{gp}||_F \leq 2\left(1 + \frac{||\Omega||}{\sigma_{min}}\right)\delta
  \label{eq:approxTheory}
\end{equation}
where $\sigma_{min}$ is the minimum singular value of $Q^{\top}\Omega$ and $\delta$ is a user defined
tolerance for how much the projection $QQ^{\top}$ distorts $\affinity$, ie:
\[
||\left(I- QQ^{\top} \right)\affinity \,|| \leq \delta.
\]

However, these results are not generally
comparable.  To the best of our knowledge no lower bounds exist
showing limits to the quality of the approximations formed by either method.
So theoretical comparisons of these bounds are ill advised if not entirely
meaningless. Furthermore, it is not clear that reconstruction of the matrix
$\affinity$ is beneficial to the machine learning tasks that the eigenvectors of $\affinity$
are used to accomplish. SCA methods often serve as
dimensionality reduction tools to create features for classification,
clustering, or regression methods. This first pass dimension reduction
does not obviate the need for doing feature
selection with the computed approximate eigenvectors.  In other words, the approximation methods
help avoid one curse of dimensionality, in that large amounts of data incurrs the cost of a large amount of
computations.  An open question is: can the selection of the tuning parameter $m$ also
avoid another curse of dimensionality in that over-parameterizing the model leads to an increase
in its prediction risk?
\begin{algorithm}[t]
\caption{\approxEig} 
Given an $n \times n$ matrix $\affinity$ and integer $m<n$, we compute an orthonormal
matrix $Q$ that approximates col$(\affinity)$ and use it to compute 
an approximation $U^{gp}$ to the  eigenvectors of $\affinity$, $U$. 
\begin{algorithmic}[1]
\STATE Draw $n\times m$ Gaussian random matrix $\Omega$.
\STATE Form $Y = \affinity \Omega$.
\STATE Construct $Q$, an orthonormal matrix such that col$(Q)$ = col$(Y)$.
\STATE Form $B$ such that $B$ minimizes  $|| BQ^{\top}\Omega - Q^{\top}Y||_2$, ie. $B$ is the least
squares solution.
\STATE Compute the eigenvector decomposition of $B$, ie: $B = \widehat U \widehat \Sigma \widehat U^{\top}$
\STATE Return $U^{gp} = Q \widehat U$.
\end{algorithmic} 
\label{alg:approx}
\end{algorithm}

A major difference in the \nys method versus the \approxEig method
is that $U^{gp}$ remains an orthonormal matrix, and hence
provides an orthogonal basis for col$(\affinity)$.  This can be seen
by considering the output of Algorithm \ref{alg:approx}, writing $u_i = U^{gp}_{N,i}$ as
the $i^{th}$ column, and observing
\begin{equation}
\langle u^{gp}_i, u^{gp}_j \rangle = \langle Q \hat{U}_{N,i}, Q \hat{U}_{N,j} \rangle = 
\langle \hat{U}_{N,i}, \hat{U}_{N,j} \rangle = \delta_{ij}
\end{equation}
by the orthonomality of $Q$ and $\hat{U}$.  This is in contrast to Algorithm \ref{alg:nystrom}
that produces $U^{nys}$ which is a rotation of $\affinity_{N,M}$ by the matrix $U^{(m)}$.  
This difference has implications that need further exploration.  However,
it does show that \approxEig provides a numerically superior set of features with which to
do learning.

\section{Experimental results}
\label{sec:experimental-results}

In order to evaluate the
approximation methods mentioned above, we focus on three related tasks:
manifold reconstruction, clustering, and classification. In the first
two cases, we examine simulated manifolds with and without an embedded
clustering problem.  For the classification task, we examine the MNIST
handwritten digit database  \citep{lecun1998}, using diffusion maps for feature creation
and support vector machines (SVMs) to
perform the classification.  

\begin{figure}[t!]
  \centering
  \includegraphics[width=3in]{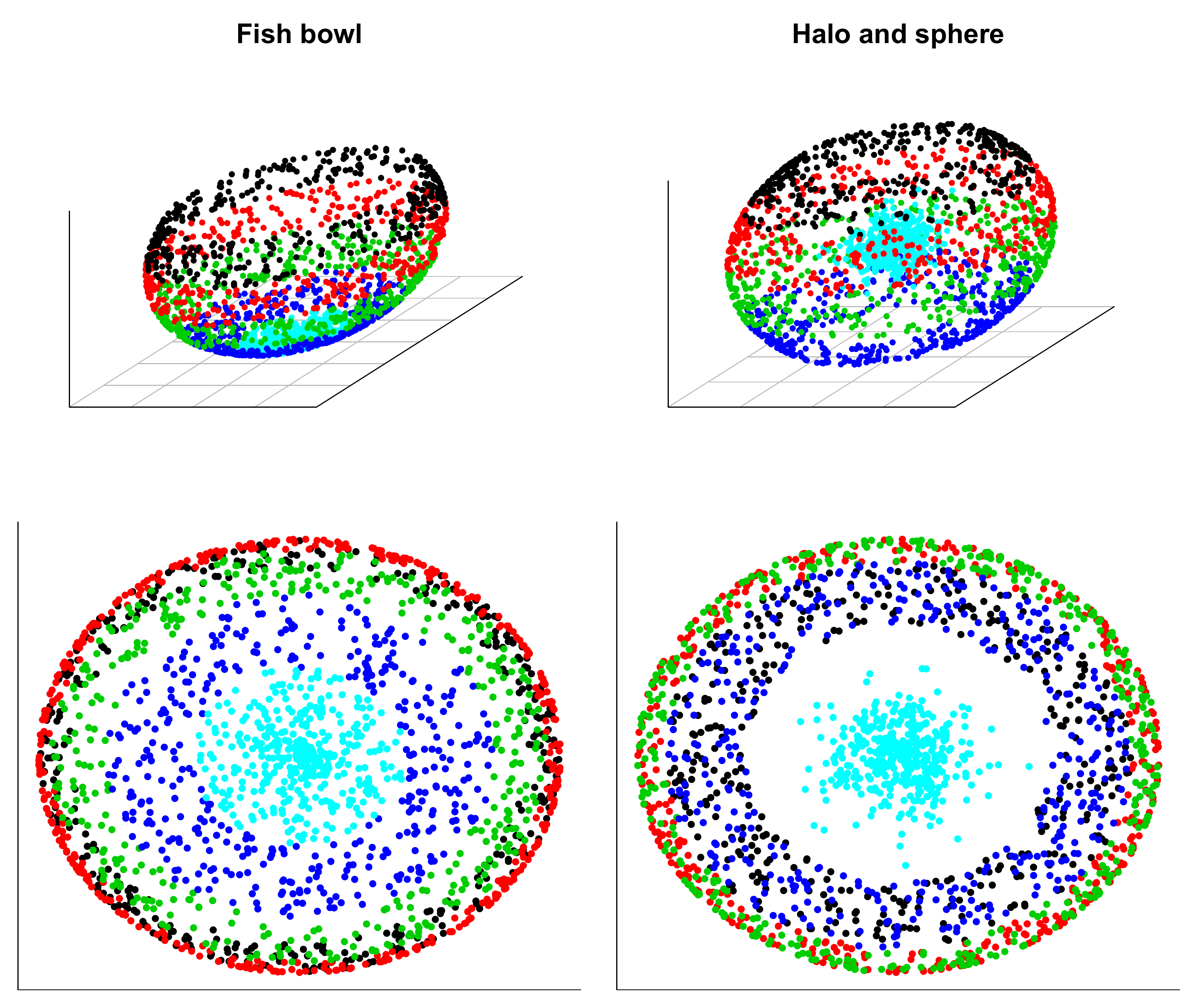}
  \caption{This figure shows the two simulated manifolds we attempt to
    recover. The fishbowl on the left is a standard exercise in the
    literature, while the modified halo and ball on the right allows
    for a related clustering exercise} 
  \label{fig:manifolds}
\end{figure}

\subsection{Manifold recovery}
\label{sec:manif-recov-clust}

To investigate the performance of the two approximation methods
through the lens of manifold recovery, we choose a standard
``fishbowl'' like object. It is constructed using the 
parametric equations for an ellipsoid:
\begin{align*}
  u &\in [0,2\pi) && x= a\cos u \sin v\\
  v &\in [d, \pi] && y = b \sin u \sin v\\
 &  && z = c \cos v,
\end{align*}
where $(a,b,c)> 0$ determine the size of the ellipsoid and $d \in [0,\pi)$ determines
the size of the opening. We sample $v$ uniformly over
the interval, while $u$ is a regularly spaced sequence of 2000
points. 

The dominant computational cost for the approximations are
$O(nm^2)$ for \nys and $O(n^2m)$ for \approxEig, we choose the parameter
$m$ so that they have equal dominant cost.  That is, if we choose $m_{gp}$
for the \approxEig method to
be a certain value, we set $m_{nys}$ for the \nys method to be $m_{nys} = \sqrt{nm_{gp}}$. For the \nys method, we take $m_{nys}=141$ while for \approxEigNoSpace, $m_{gp}=10$.

\begin{figure}[t!]
  \centering
  \includegraphics[width=5.5in]{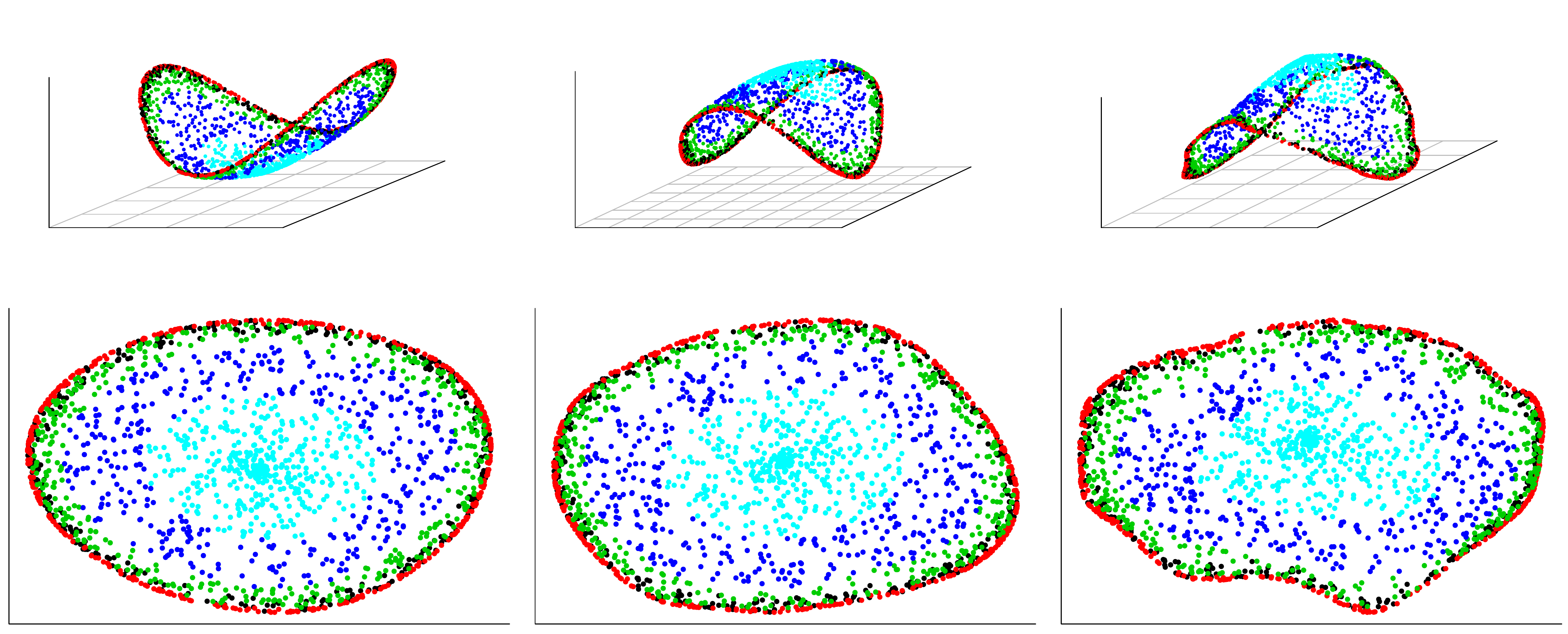}
  \caption{Here, the true eigenvectors along with both approximate methods
    demonstrate the manifold recovery capabilities. For equivalent
    computational time, both methods perform reasonably. The true
    eigenvectors are on the left, the \nys method is in the center,
    while the \approxEig method is on the right.} 
  \label{fig:bowl}
\end{figure}

The manifold reconstruction of the ellipsoid is shown in
Figure~\ref{fig:bowl}. Both methods perform reasonably well. The
underlying structure is well recovered. For this figure, the tuning
parameter was set at $\epsilon=15$ in all three cases. Changing it
makes very little difference ($\epsilon \in [1,200]$ yields very
similar figures over repeated randomizations\footnote{The magnitude of
  the tuning parameter depends heavily on the intrinsic separation in
  the data}). This robustness with
respect to tuning parameter choice is rarely the
case as we will see in the next example.\footnote{Another standard manifold recovery example, the
  ``swissroll'', is very sensitive to tuning parameter choice as well
  as the choice of the matrix $\widetilde{\affinity}$.}

\subsection{Clustering}
\label{sec:clustering}

\begin{figure}[t!]
  \centering
  \includegraphics[width=5.5in]{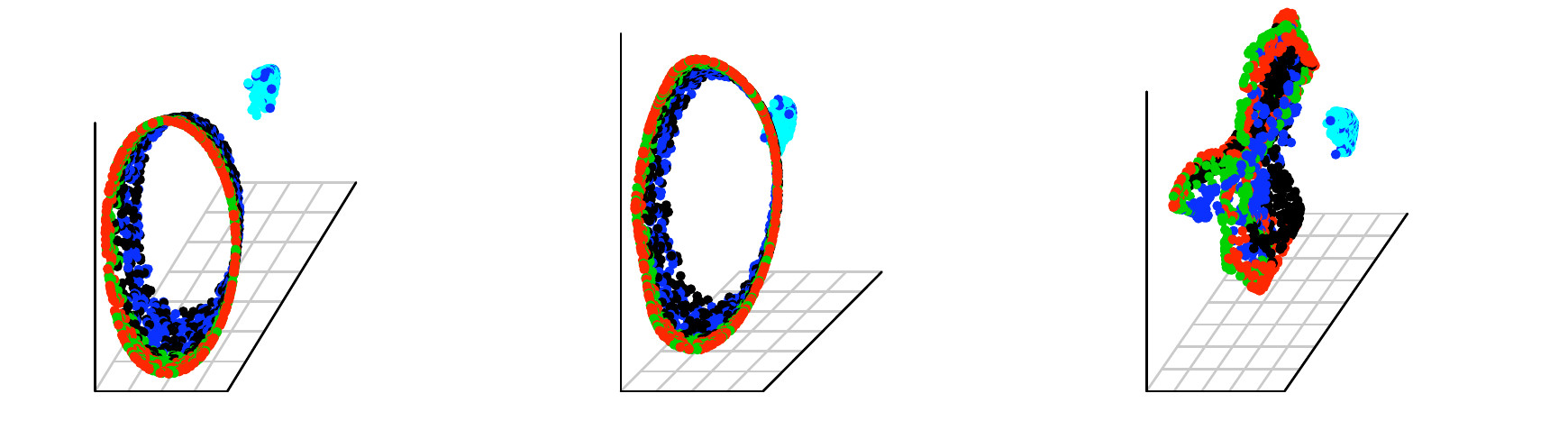}\\
  \includegraphics[height=2in,width=5.5in]{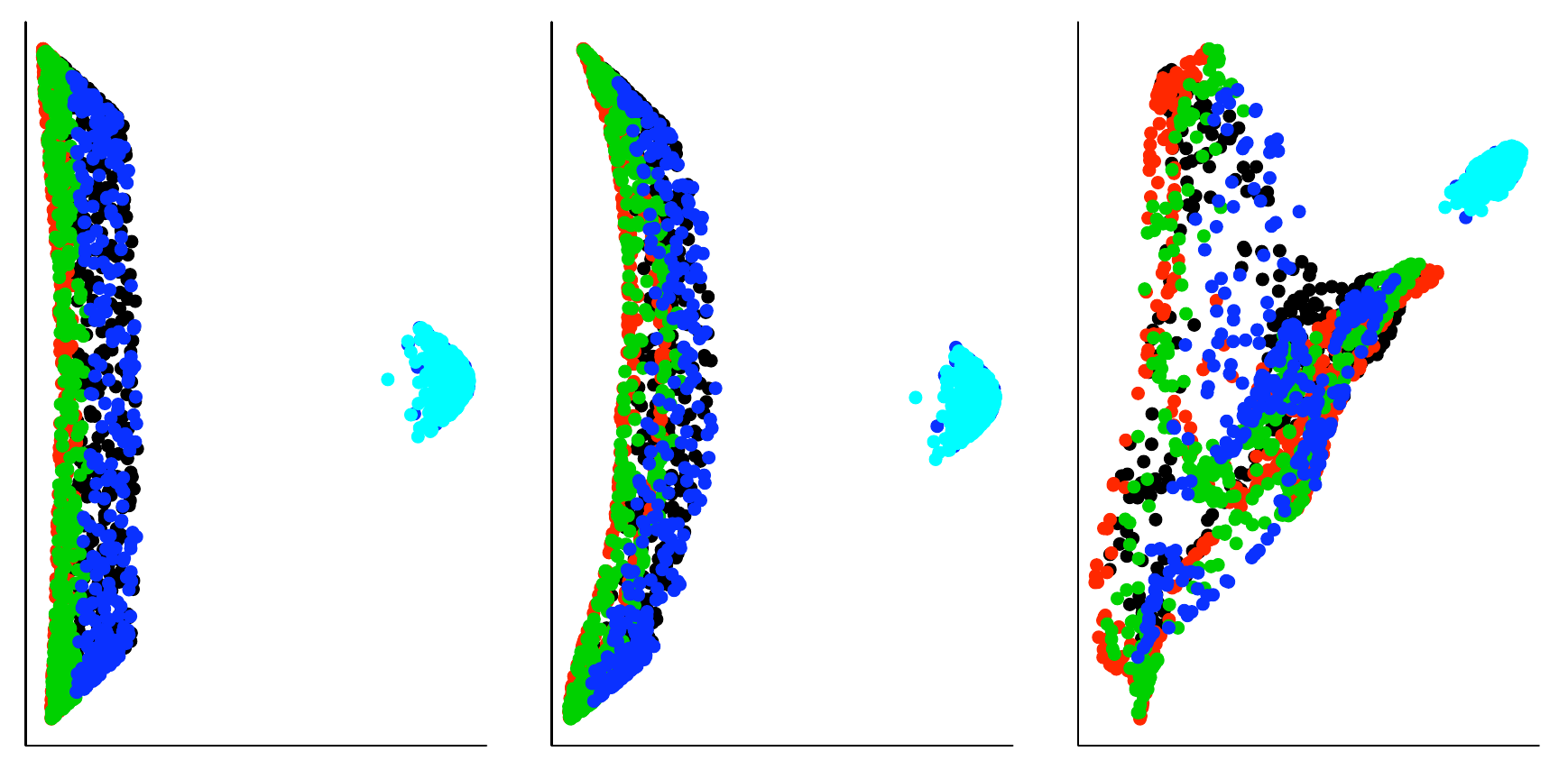}
  \caption{Using the true eigenvectors and with both approximation
    methods, we
    attempt to perform a clustering task. For equivalent computational
    time, the \nys method far outperforms \approxEigNoSpace. The true
    eigenvectors are on the left, the \nys method is in the center,
    while the \approxEig method is on the right.}
  \label{fig:halo}
\end{figure}

To create a clustering example similar to the manifold recovery
problem, we removed the bottom fifth of the fishbowl and created a glob in
the middle of the resulting halo.  See Figure~\ref{fig:manifolds} for a plot of the shape. 
We consider the light blue observations in the middle one cluster 
and the outer ring as a second cluster.
The result is a three dimensional
clustering problem for which linear classifiers will fail. However,
diffusion methods yield an embedding which will be separable even in
one dimension via linear methods (Figure~\ref{fig:halo}, first column). 
The next two columns show the
\nys approximation with $m_{nys} = 200$ and for \approxEig with $m_{gp}=20$.
We see that \nys provides a very faithful reproduction of the exact eigenvectors.
\approxEig loses much of the separation present in the other two methods.

It is important to keep in mind the selection of appropriate tuning parameters.
Our choice in this example is a subjective one, based on visual inspection.
In this scenario, tuning parameter selection becomes very difficult,
and hugely important. Small perturbations in the tuning parameter lead
to poor embeddings which not only yield poor clustering solutions, but
which can completely destroy the structure visible in the data
alone. Furthermore, tuning parameters must be selected separately for
the exact as well as the two approximate methods with no guarantee
that they will be similar. In our particular parameterization, the
exact reconstruction was successful for $\epsilon \in [0.05,0.15]$ while
the approximate methods required $\epsilon\approx 0.25$. The
reconstruction via \approxEig shown in Figure~\ref{fig:halo}
displays the best separation that we could achieve for this
computational complexity. Clearly, the \nys method performs better in
this case, yielding an embedding that remains easily separable even in
one dimension.

\subsection{Classification}
A common machine learning task is to classify data based on a labelled set
of training data.  When all the data (training and test) are available,
using the semi-supervised approach of a diffusion map from the graph
Laplacian is a very reasonable technique for providing labels for
the test data.  However, this technique requires a very large spectral decomposition
as it is based on the entire dataset, both training and test.

\begin{figure}[t!]
  \centering
  \includegraphics[width=2.65in]{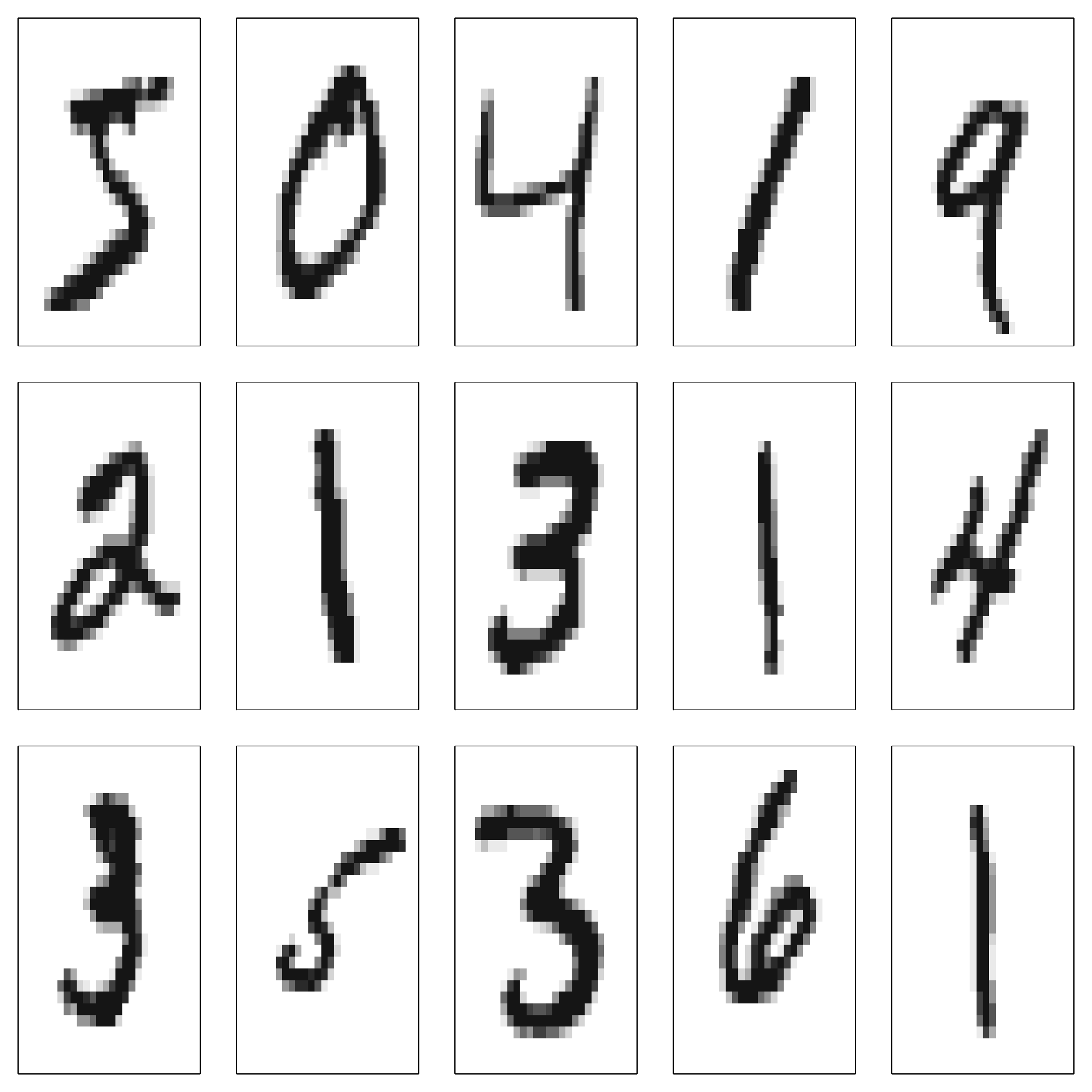}
  \includegraphics[width=2.65in]{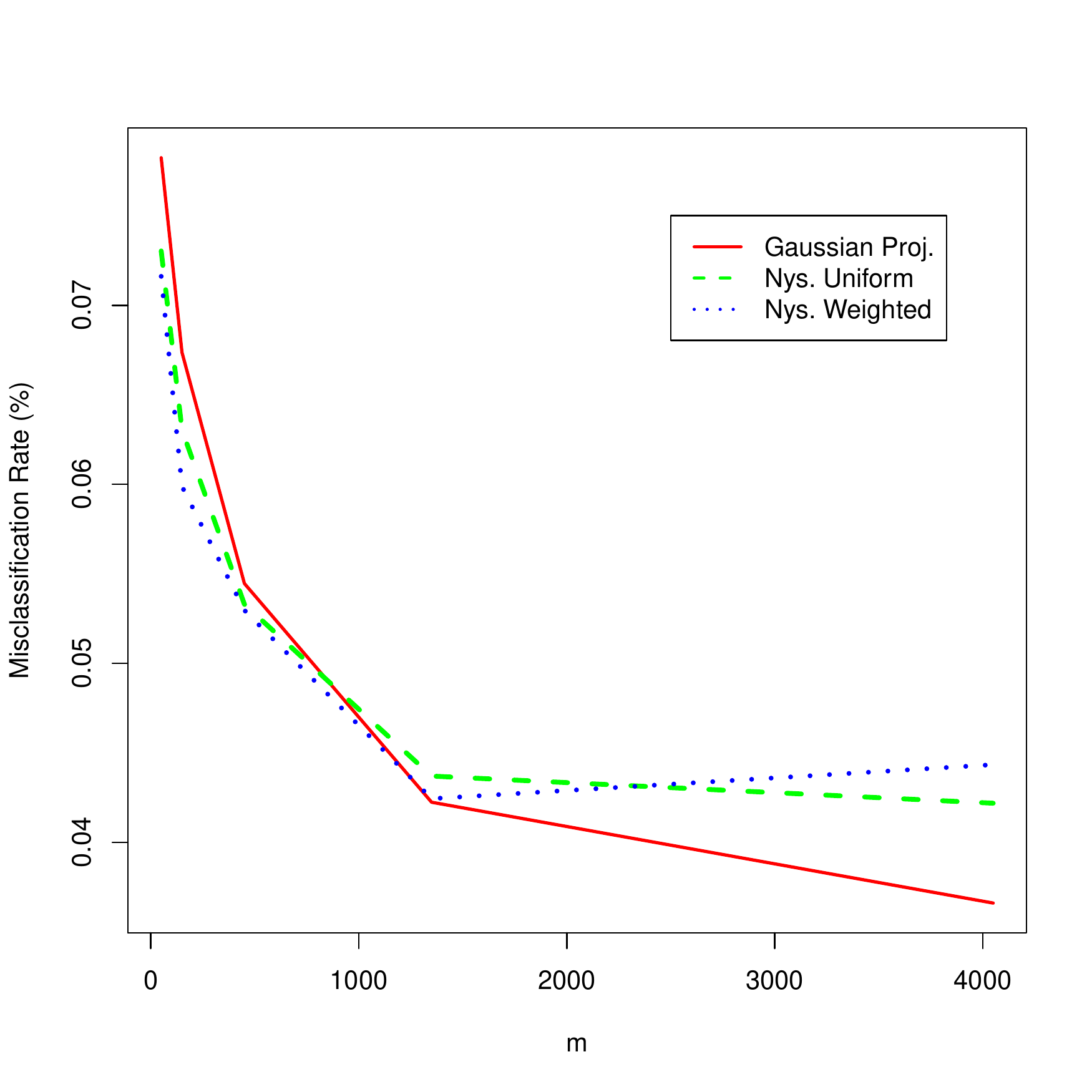}
  \caption{Left: A random sample of fifteen digits from the MNIST dataset.  
           Right: Misclassification rates from running the three approximation methods on a random
           subset of the MNIST digits for $m \in  \{ 50,\ 150,\ 450,\ 1350,\ 4050\}$ with 
           $\ntrain = 15000$ and $\ntest = 3000$ (see text for more details).  
           The weighted and uniform \nys methods are very similar.  The \approxEig method
           results in a worse misclassification rate for smaller $m$, but its performance
           improves markedly for larger $m$.}  
  \label{fig:handwritten}
\end{figure}

We investigate classifying the handwritten digits found in the MNIST dataset
and used in e.g. \cite{lecun1998}.  See the left panel of 
Figure \ref{fig:handwritten} for fifteen randomly selected
digits.  Using SVM with the
(approximate) eigenvectors of the graph Laplacian as features, we 
attempt to classify a test set using each
of the above described approximation techniques.  Though the digits are
rotated and skewed relative to each other, we do not consider 
deskewing as very good classification results
have been obtained using
SVM without any such preprocessing (see for example~\citep{lauer2007}).  We choose the smoothing
parameter in the SVM via 10 fold cross-validation.  Additionally,
we choose the bandwidth parameter of the diffusion map, $\epsilon$,
by minimizing the misclassification error on
the test set over a grid of values.

For small enough datasets, we can compute the true eigenvectors to get an idea of the efficiency
loss we incurr by using each approximation.  If we choose a random dataset comprised of 
$\ntrain = 4000$ test digits and $\ntest = 800$ training digits, 
then it is still feasible to calculate the true eigenvectors of the matrix $\affinity$.
Table~\ref{tab:m400} displays the results with an approximation
parameter of $m=m_{nys}=m_{gp}=400$.\footnote{In the manifold recovery
  and clustering tasks, we are essentially interested only in the
  first few eigenvectors of $\affinity$, so it makes sense to compare
  the two methods for equivalent computational complexity. However in
  the classification task, we want to create two comparable feature
  matrices which may then be regularized by the classifier. For this
  reason, it seems more natural to choose $m_{nys}=m_{gp}$.}
\begin{table}[tbp]
   \centering
   \begin{tabular}{@{} lr @{}}
      \hline
       \textbf{Method}        & \# Correct \\
       \hline
      $\ntest$                & 800 \\
      True eigenvectors       & 756 \\
      Uniform \nys            & 697 \\
      Weighted \nys           & 701 \\      
       \approxEig             & 725 \\      
       \hline
   \end{tabular}
   \caption{$\ntrain = 4000$, $m = 400$}
   \label{tab:m400}
\end{table}
We see that the \approxEig method has the best performance of the three considered
methods.  Additionally, as expected, the weighted \nys method outperforms 
the uniform \nysNoSpace, but only slightly.  Note that the true eigenvector misclassification rate, 5.5\%,
is a bit worse than that reported in \cite{lauer2007} --- 1\%.  We attribute this
to using a much smaller training sample for our classification procedure. 

Additionally, we ran a comparison with $\ntrain = 15000$ and $\ntest = 3000$.  
Here computing the true eigenvectors would begin
to be truly time/resource consuming.  For each $m \in \{ 50,\ 150,\
450,\ 1350,\ 4050\}$ 
we perform the grid search
to minimize test set misclassification for each method.  We do this a total of ten
times and average the results to somewhat account for the randomness of each approximation method.
The results appear on the right hand side of Figure \ref{fig:handwritten}.
We see that again the weighted and uniform \nys methods produce very similar misclassification rates.  
Interestingly, the \approxEig method
results in a worse misclassification rate for smaller $m$, but its performance
improves markedly for larger $m$.

\section{Discussion}
\label{sec:discussion}
In this paper, we consider two main methods --- the \nys extension 
and another method we dub \approxEig --- for approximating the eigenvectors of
the graph Laplacian matrix $\affinity$.  As a metric for performance, we are interested
in ramifications of these approximations in practice.  That is, how do the approximations
affect the efficacy of standard machine learning techniques in standard machine learning tasks.
  
Specifically, we consider three common applications: manifold reconstruction, clustering,
and classification.  For the last task we additionally investigate a newer version
of the \nys method, one that allows for weighted sampling of the columns of $\affinity$
based on the size of the diagonal elements.  

We find that for these sorts of applications, neither \nys nor
\approxEig achieves uniformly better results. For similar computational costs,
both methods perform well at manifold reconstruction, each finding 
the lower dimensional structure of the fishbowl as a deformed plane.  However, the \approxEig
method recovery is somewhat distorted relative to the true eigenvectors. In the 
clustering task, \approxEig
performs relatively poorly.  Although the inner sphere is now linearly separable from the outer ring,
the embedding is much noisier than for the true eigenvectors and the \nys extension.
Lastly, we find that the weighted and uniform \nys methods result in nearly the same misclassification
rates, with perhaps a slight edge to the weighted version.  But, \approxEig outperforms both
\nys methods as long as $m$ is moderately large: about 10\% of $n$.  This value of $m$ still provides 
a substantial savings over computing the true eigenvectors.


\bibliography{nystromNotCrazy}
\bibliographystyle{mybibsty}
\end{document}